\documentclass[conference]{IEEEtran}
\usepackage[a4paper, top=19mm, bottom=43mm, left=14mm, right=14mm]{geometry}

\usepackage{cite}
\usepackage{amsmath,amssymb,amsfonts}
\usepackage{algorithmic}
\usepackage{graphicx}
\usepackage{textcomp}
\usepackage{xcolor}
\def\BibTeX{{\rm B\kern-.05em{\sc i\kern-.025em b}\kern-.08em
    T\kern-.1667em\lower.7ex\hbox{E}\kern-.125emX}}

\usepackage{subcaption}
\usepackage{amsfonts}
\usepackage{multirow}
\usepackage{enumitem}
\newcommand{\reffig}[1]{figure \ref{#1}}

\begin{document}

\title{
  A Privacy-Preserving \\
  Semantic-Segmentation Method \\
  Using Domain-Adaptation Technique
% {\footnotesize \textsuperscript{*}Note: Sub-titles are not captured for https://ieeexplore.ieee.org  and
\thanks{Identify applicable funding agency here. If none, delete this.}
}

\author{\IEEEauthorblockN{1\textsuperscript{st} Homare Sueyoshi}
\IEEEauthorblockA{\textit{Tokyo Metropolitan University} \\
% \textit{Tokyo Metropolitan University}\\
6-6, Asahigaoka, Hino, Tokyo, Japan \\
sueyoshi-homare@ed.tmu.ac.jp}
\and
\IEEEauthorblockN{2\textsuperscript{nd} Kiyoshi Nishikawa}
\IEEEauthorblockA{\textit{Tokyo Metropolitan University} \\
% \textit{Tokyo Metropolitan University}\\
6-6, Asahigaoka, Hino, Tokyo, Japan \\
kiyoshi@tmu.ac.jp}
\and
\IEEEauthorblockN{3\textsuperscript{rd} Hitoshi Kiya}
\IEEEauthorblockA{\textit{Tokyo Metropolitan University} \\
% \textit{Tokyo Metropolitan University}\\
6-6, Asahigaoka, Hino, Tokyo, Japan \\
kiya@tmu.ac.jp}
% \and
% \and
% \and
}

\maketitle

\begin{abstract}
We propose a privacy-preserving semantic-segmentation method for applying perceptual encryption to images used for model training in addition to test images.
This method also provides almost the same accuracy as models without any encryption.
The above performance is achieved using a domain-adaptation technique on the embedding structure of the Vision Transformer (ViT).
The effectiveness of the proposed method was experimentally confirmed in terms of the accuracy of semantic segmentation when using a powerful semantic-segmentation model with ViT called Segmentation Transformer.
\end{abstract}

\begin{IEEEkeywords}
Semantic Segmentation, Vision Transformer, Image Encryption, Privacy-Preserving
\end{IEEEkeywords}

\section{Introduction}
Deep-learning models typically require large datasets and huge computational resources for training, which often leads to the use of cloud servers.
There is also a trend of deploying trained models on cloud servers to provide inference services.
However, data, such as images, generally contain sensitive personal information, and cloud servers are not always completely trustworthy \cite{sirichotedumrong2018grayscale, kiya2022overview}.
Therefore, it is important to consider privacy protection when using cloud servers.
We propose a method for privacy-preserving semantic segmentation using Segmentation Transformer (SETR) \cite{Zheng_2021_CVPR} that is based on Vision Transformer (ViT)\cite{dosovitskiy2021an}.
The proposed method applies perceptual encryption to both training and testing images, aiming to protect visual information throughout the learning and inference process.

\section{Related work}
In cloud-based environments, research on model training with perceptual encryption began in image classification \cite{kiya2023image}.
Early studies mainly focused on convolutional-neural-network models, but a key issue was performance degradation caused by the use of encrypted images.
This issue was significantly alleviated by the discovery that isotropic networks, such as ViT and ConvMixer, which are based on patch-wise processing, are well-suited for block-wise image encryption \cite{kiya2022overview, 9760030, lin2024privacy, kiya2023image}.
The use of encrypted data has also been effective in adversarial defense \cite{iijima2024random, aprilpyone2021block} and access control \cite{maungmaung2021protection} in addition to in privacy preservation \cite{kiya2022overview, iida2020privacy, nagamori2024efficient}.

In semantic segmentation, it has been shown that using ViT-based models can achieve accuracy comparable to that with non-encrypted images, even when using encrypted images\cite{kiya2022privacy}.
As illustrated in Figure \ref{fig:setr_test_encryption}, in the framework of the model provider first trains the model on non-encrypted images.
Encryption is then applied to certain parameters of the trained model.
The encrypted model is uploaded to a cloud server, and the encryption key used is shared with end users.
Users encrypt their test images using the shared key, and these encrypted images are processed in the cloud while visual information remains protected.
Importantly, neither the key nor the detailed visual content of the images is exposed to the cloud server.

\begin{figure}[tb]
  \centering
  \includegraphics[width=\linewidth]{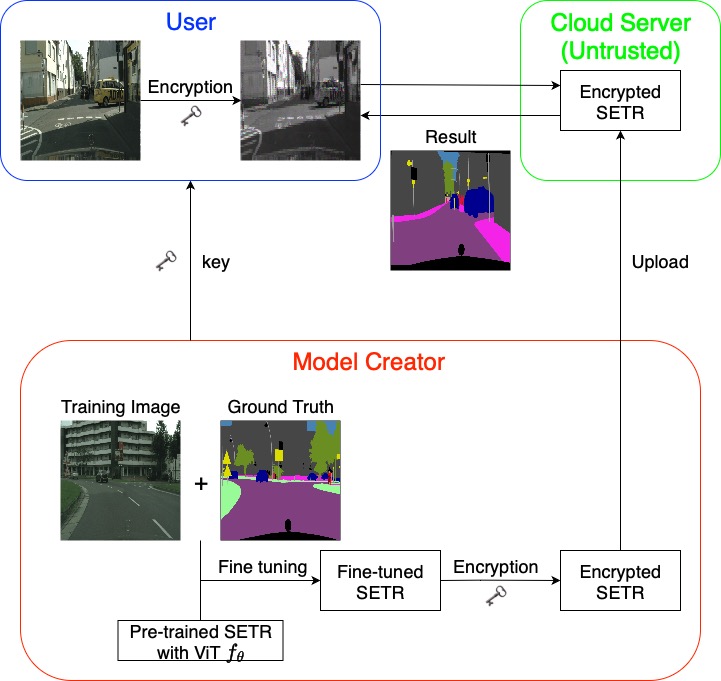}
  \caption{Framework of conventional method 1 \cite{kiya2022privacy}}
  \label{fig:conventional_method}
\end{figure}

Since cloud servers are not necessarily trustworthy, this approach requires that model training be executed locally using non-encrypted images.
To enable secure execution on cloud servers, therefore, previous methods must compromise model accuracy \reffig{fig:conventional_method}.
The proposed method enables training of semantic-segmentation models directly in cloud environments while achieving accuracy equivalent to that of models trained and tested with non-encrypted data, as in prior work.

\section{Proposed Method}
\subsection{Overview of proposed method}
\begin{figure}[tb]
  \centering
  \includegraphics[width=\linewidth]{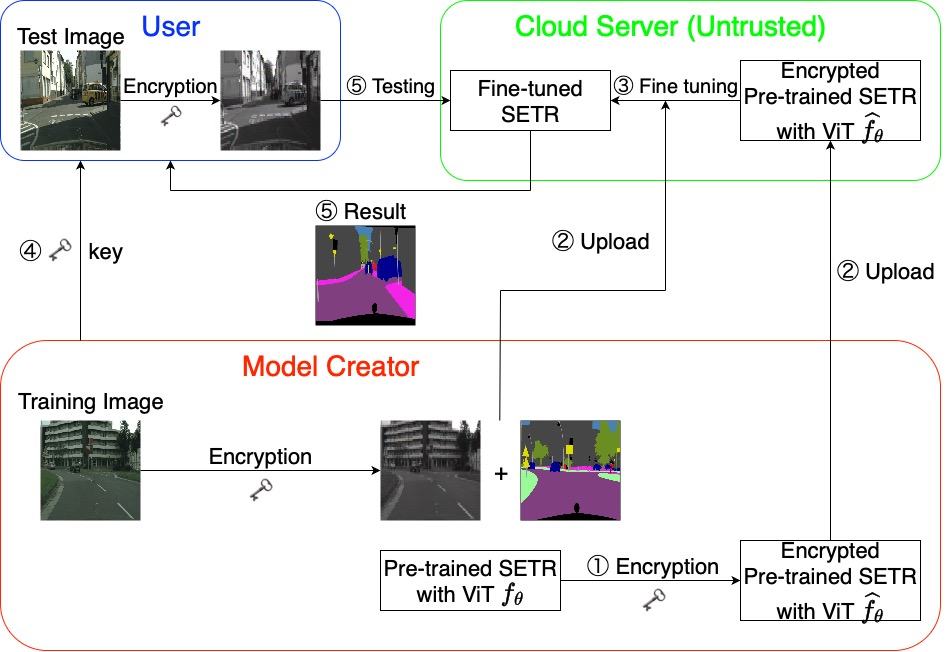}
  \caption{Framework of proposed method}
  \label{fig:proposed_method}
\end{figure}

\begin{figure}[tb]
  \centering
  \includegraphics[width=\linewidth]{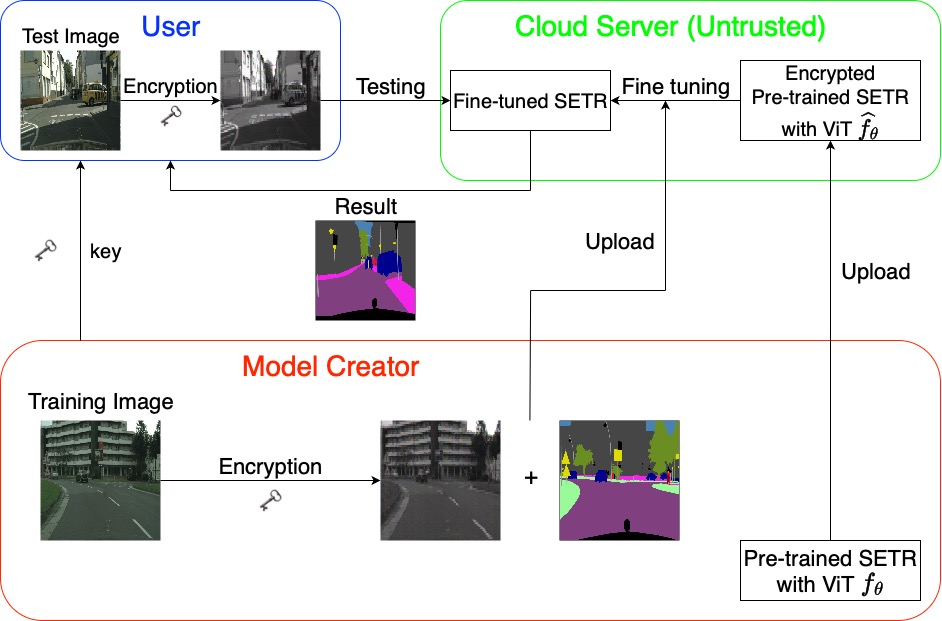}
  \caption{Framework of conventional method 2}
  \label{fig:setr_test_encryption}
\end{figure}

The framework of the proposed method is summarized below (see \reffig{fig:proposed_method}).

\begin{enumerate}[label=\textcircled{\raisebox{-0.7pt}{\small\arabic*}}]
  \item \label{step:encrypt}
  The model creator generates random sequences using a secret key for each patch and encrypts the patch embedding of the pre-trained model and training images by using the sequences.

  \item
  The encrypted pre-trained model $\hat{f_\theta}$ with the ground-truth labels of the segmentation map and encrypted training images are uploaded to the cloud server.

  \item
  The $\hat{f_\theta}$ is fine-tuned to construct an encrypted model by using the encrypted training images on the cloud server.

  \item
  The user receives the secret key from the model creator, encrypts a query image, and uploads them to the cloud server.

  \item
  The encrypted test image is input to the fine-tuned model, then an estimated result is sent out to the user.
\end{enumerate}

In the above framework, the cloud server does not have plain images and the secret key.
The main difference between the conventional and proposed methods is that pre-train models before fine-tuning the models, as shown in \reffig{fig:proposed_method}.

\subsection{Image Encryption}
Encrypted images are used for fine-tuning models and testing.
With the proposed method, a block-wise encryption method is carried out in a similar manner as with conventional methods \cite{kiya2022privacy}.
The encryption procedure is summarized as follows.

\begin{figure}[tb]
  \centering
  \captionsetup[subfigure]{labelformat=empty}
  \begin{minipage}[b]{0.48\linewidth}
    \centering
    \includegraphics[width=0.95\textwidth]{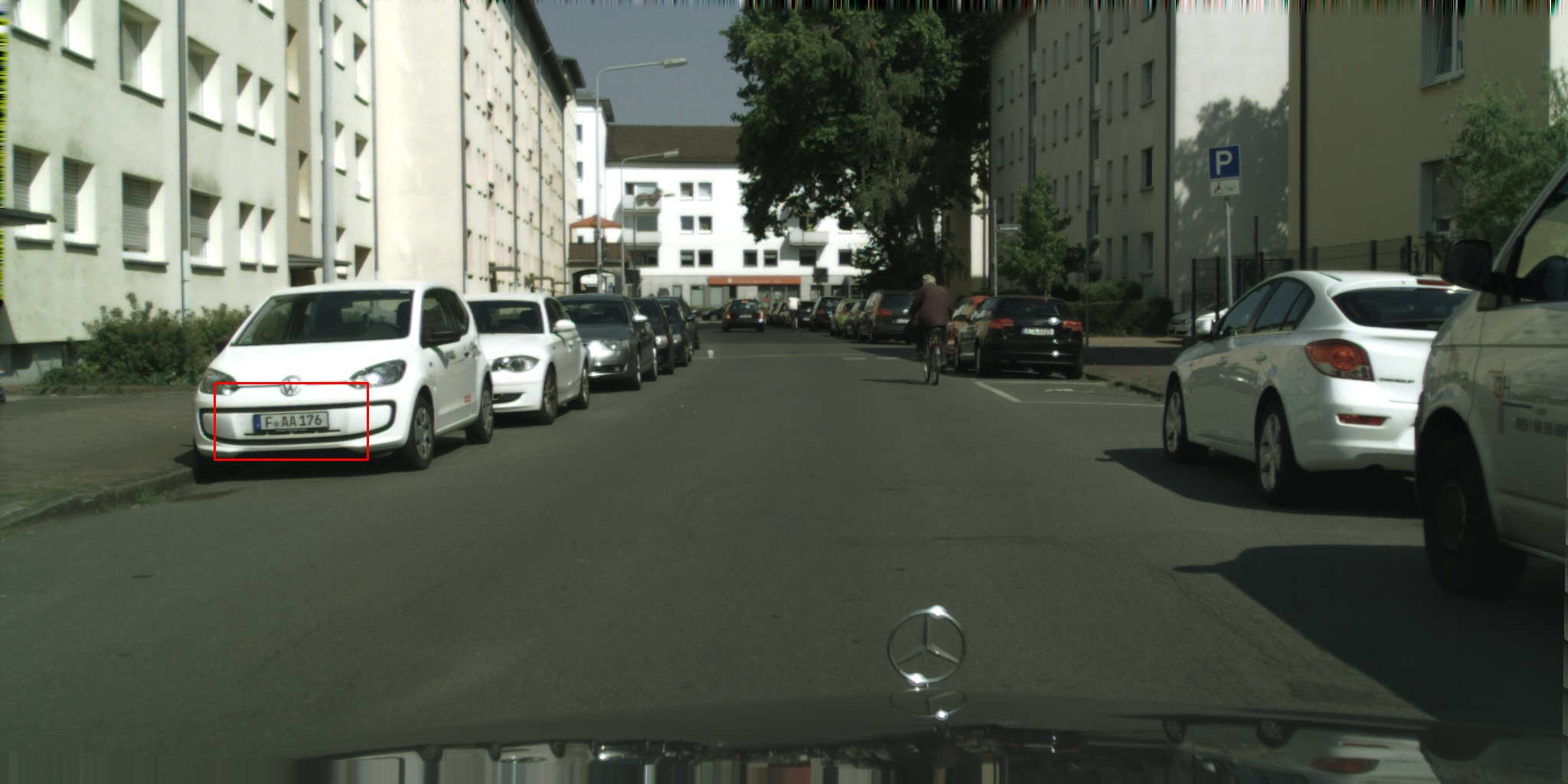}
    \subcaption{Original image}
    \label{fig:cityscapes_plain}
  \end{minipage}
  \begin{minipage}[b]{0.48\linewidth}
    \centering
    \includegraphics[width=0.95\textwidth]{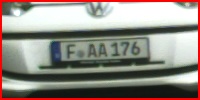}
    \subcaption{Zoomed-in view}
    \label{fig:cityscapes_plain_crop}
  \end{minipage}
  \caption*{(a) Non-encrypted image}

  \vspace{0.5em}

  \begin{minipage}[b]{0.48\linewidth}
    \centering
    \includegraphics[width=0.95\textwidth]{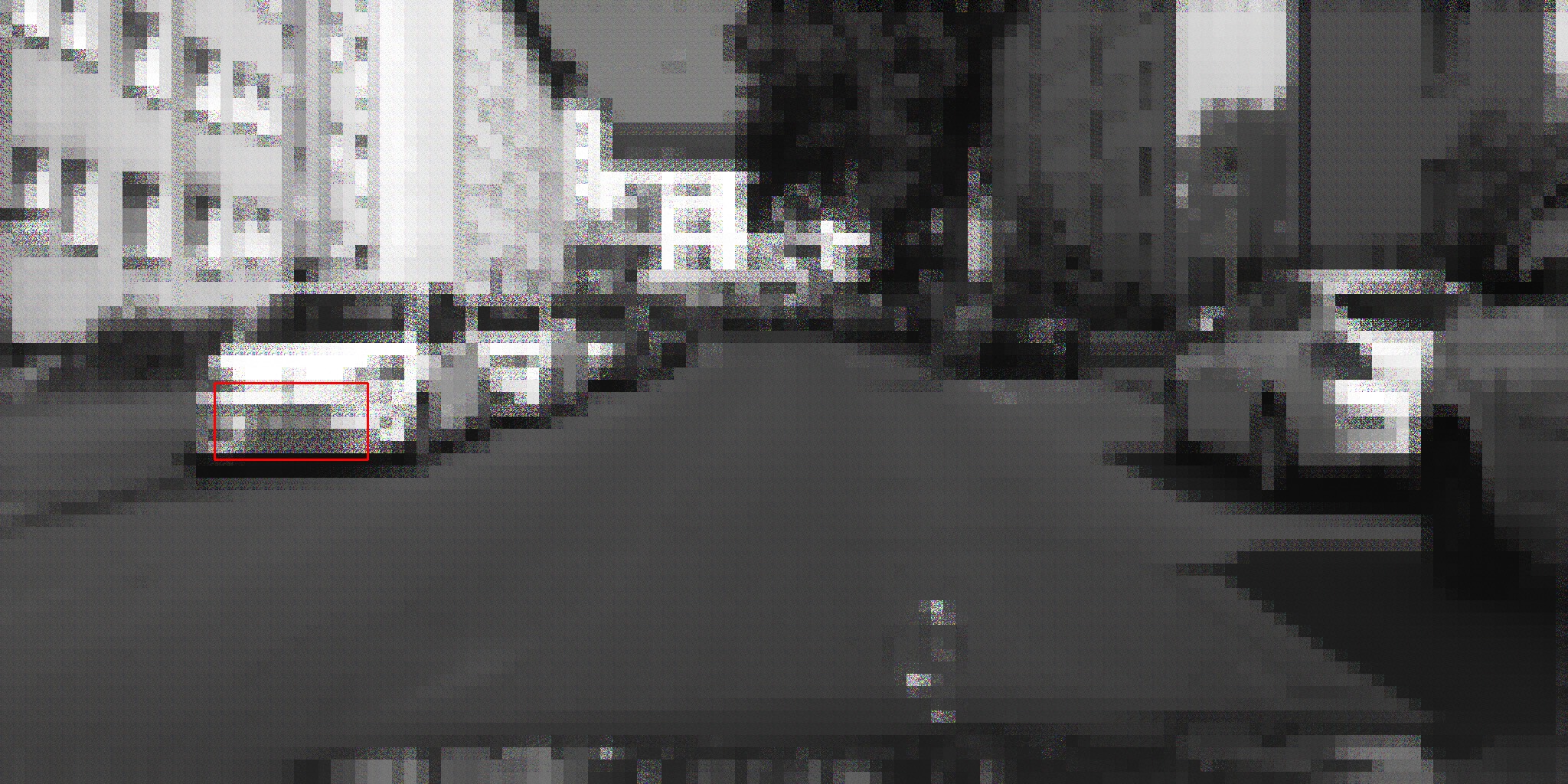}
    \subcaption{Encrypted image}
    \label{fig:cityscapes_enc}
  \end{minipage}
  \begin{minipage}[b]{0.48\linewidth}
    \centering
    \includegraphics[width=0.95\textwidth]{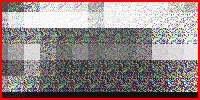}
    \subcaption{Zoomed-in view}
    \label{fig:cityscapes_enc_crop}
  \end{minipage}
  \caption*{(b) Encrypted image (proposed)}

  \caption{Example of encrypted image}
  \label{fig:cityscapes_comparison}
\end{figure}

\begin{enumerate}[label=step\arabic*:, leftmargin=3em, ref=\arabic*:]
  \item
  Divide an image into non-overlapping blocks with the same size as the patch size of the pre-trained ViT.
  \item
  Randomly shuffle pixels across three color channels within each block by using a random sequence generated with the secret key.
\end{enumerate}
Figure \ref{fig:cityscapes_comparison} shows an example of encrypted images.

\subsection{Model Encryption}
The pre-trained model is encrypted as shown in step \ref{step:encrypt}.
The patch embedding of the pre-trained model is encrypted to reduce the impact of the block-wise image encryption by using the secret key used for image encryption as in a previous study\cite{nagamori2024efficient}.

\section{Experiments}
\subsection{Setup}
We conducted semantic-segmentation experiments on the Cityscapes dataset.
The dataset consists of 5000 images with a resolution of 1024×2048 where the dataset with 19 classes includes 2975 training images, 500 validation images, and 1525 test images.
In the experiments, we used mmsegmentation from OpenMMLab and SETR.
We used the ViT pre-trained models ``vit\_base\_patch\_16\_384'' and ``vit\_large\_patch16\_384''
from the timmlibrary as the encoder's pre-trained model and MLA (Multi-Level feature Aggregation), PUP (Progressive UPsampling), and Naïve (Naive upsampling) as the decoder for fine-tuning the pre-train models.

\subsection{Experimental Results}
\label{subsec:experiment_result}
Table \ref{tab:result_cityscapes} lists the evaluation results of estimated semantic-segmentation maps estimated under various conditions involving three evaluation metrics to assess accuracy:
Acc (average accuracy), mAcc (mean accuracy), and mIoU (mean Intersection over Union), which measures the average overlap between predicted labels and ground-truth labels for each class.
mIoU is the average of IoU values for each class, where IoU is expressed as:
\begin{gather*}
  \mathrm{IoU}=\frac{TP}{TP + FP + FN}
\end{gather*}
For a given class $x$, a true positive ($TP$) represents the number of pixels correctly predicted as class $x$,
false positive ($FP$) represents the number of pixels incorrectly predicted as class $x$,
and false negative ($FN$) represents the number of pixels that belong to class $x$ but were not correctly predicted.
All metrics range from 0 to 100, with higher values indicating better performance.

\begin{figure}[tb]
  \centering
  \captionsetup[subfigure]{labelformat=empty}
  \begin{minipage}[b]{0.48\linewidth}
    \centering
    \includegraphics[width=0.95\textwidth]{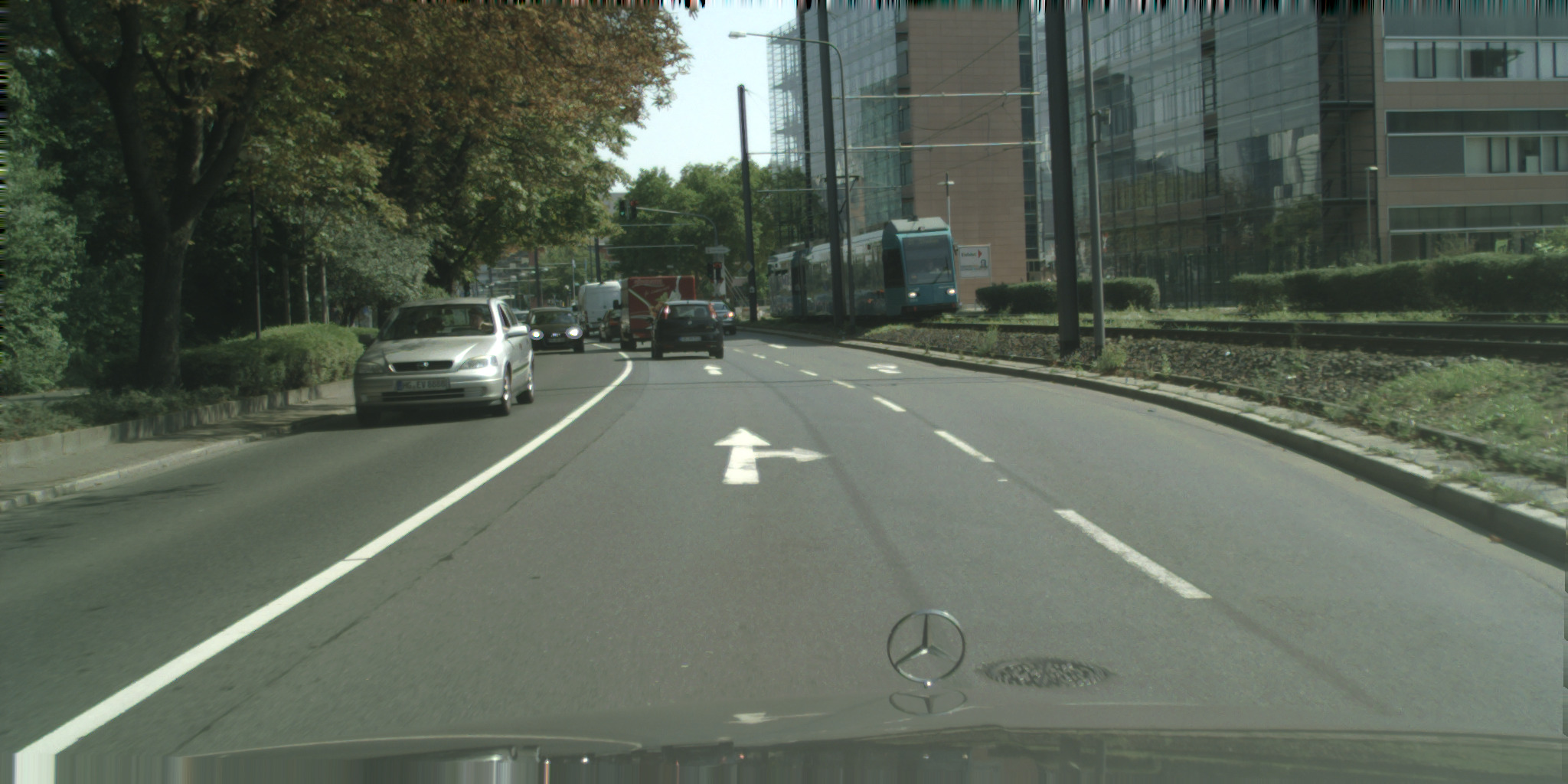}
    \subcaption{Original image}
    \label{fig:original_image}
  \end{minipage}
  \begin{minipage}[b]{0.48\linewidth}
    \centering
    \includegraphics[width=0.95\textwidth]{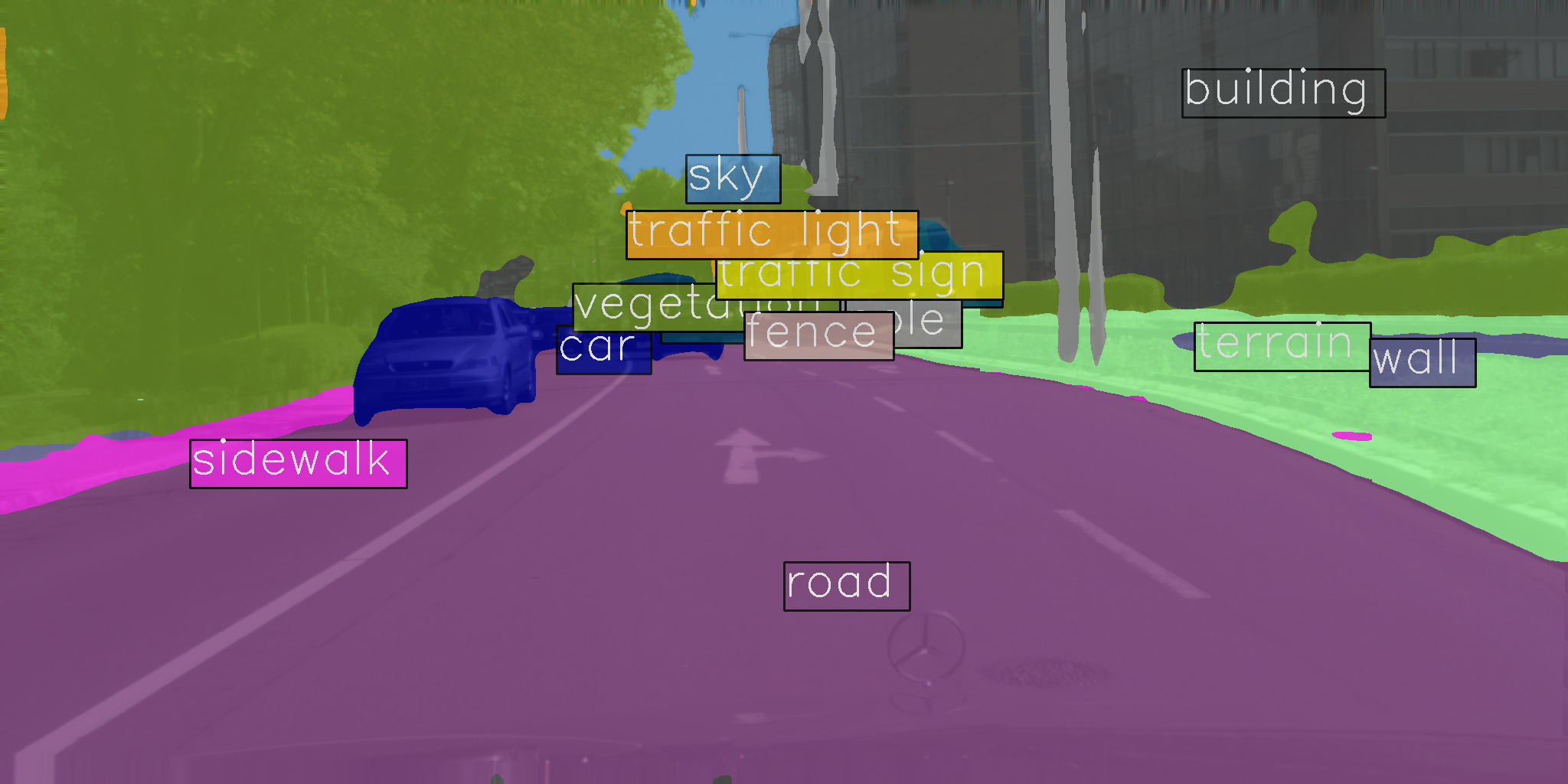}
    \subcaption{Baseline (mIoU=73.26)}
    \label{fig:cityscapes_mla_l_pm_pi_predict}
  \end{minipage}
  \caption*{(a) Non-encrypted images}

  \vspace{1em}

  \begin{minipage}[b]{0.48\linewidth}
    \centering
    \includegraphics[width=0.95\textwidth]{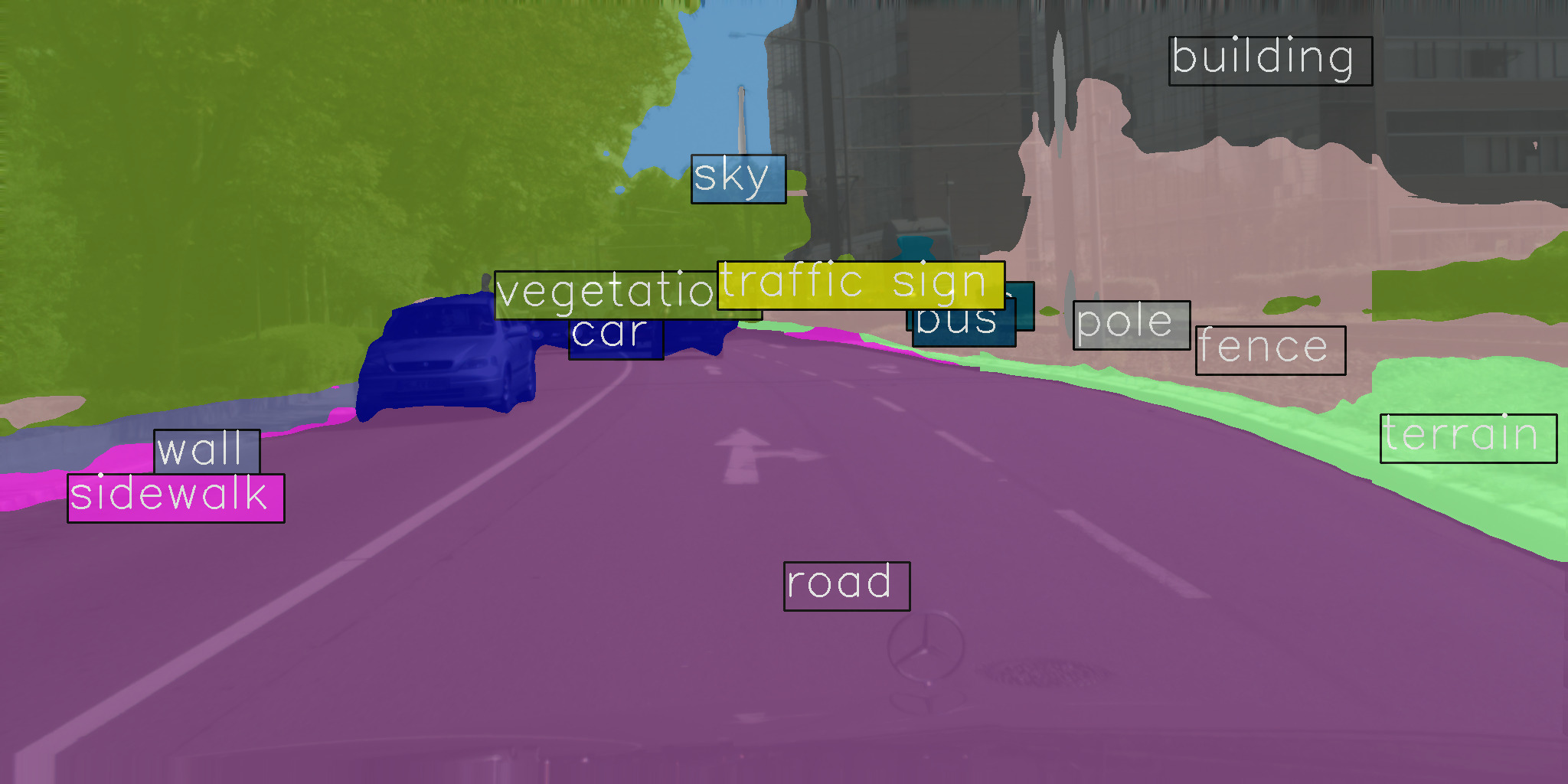}
    \subcaption{Conventional 2(mIoU=53.55)}
    \label{fig:cityscapes_mla_l_pm_ei_predict_incorrect_key}
  \end{minipage}
  \begin{minipage}[b]{0.48\linewidth}
    \centering
    \includegraphics[width=0.95\textwidth]{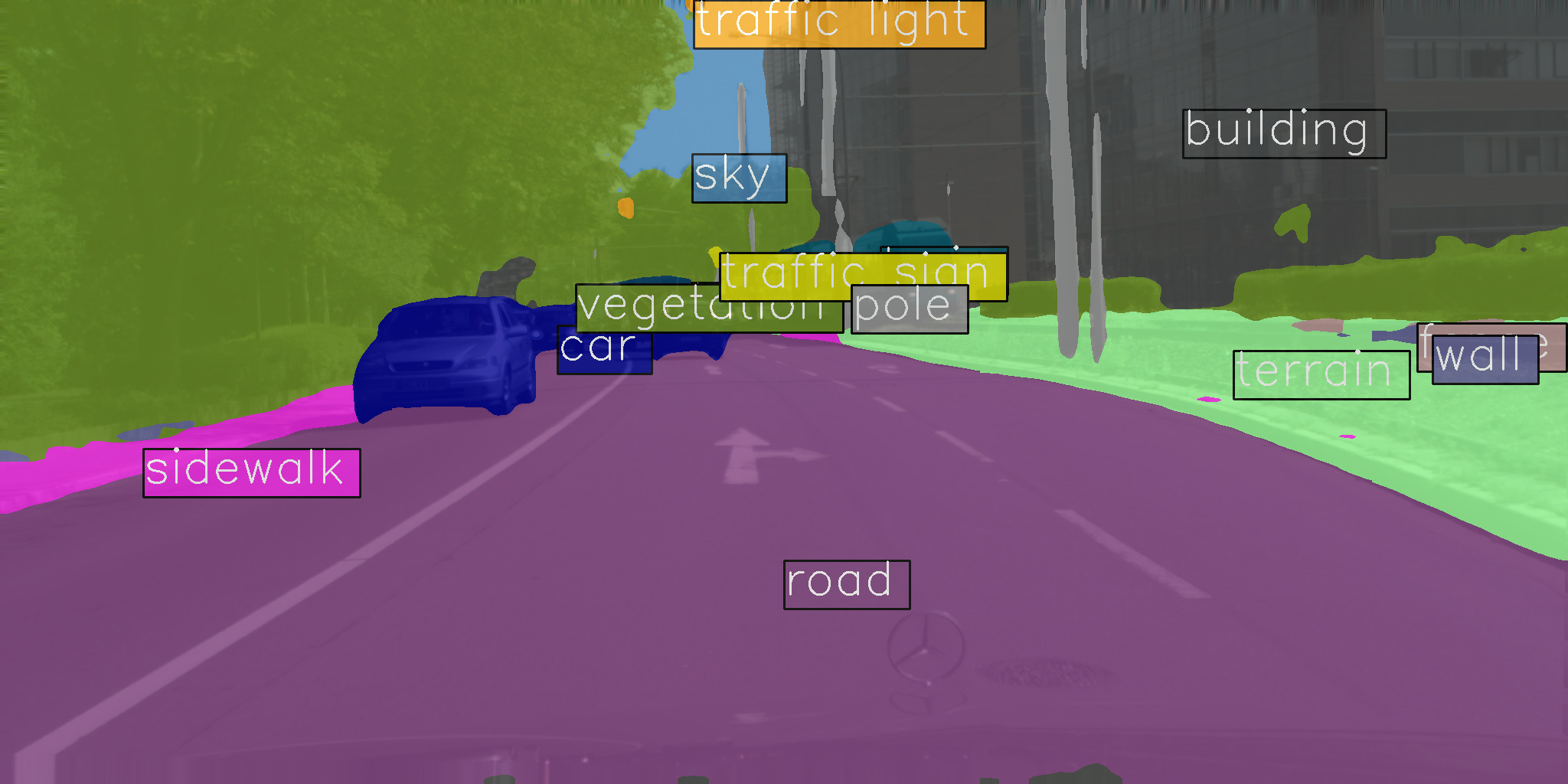}
    \subcaption{Proposed (mIoU=72.57)}
    \label{fig:cityscapes_mla_l_em_ei_predict_correct_key}
  \end{minipage}
  \caption*{(b) Encrypted images}

  \caption{Examples of segmentation maps (ViT-Large, MLA)}
  \label{fig:cityscapes_mla_l}
\end{figure}

Figure \ref{fig:cityscapes_mla_l} also shows an example output of segmentation maps for each methods
when we used ``ViT-Large'' as the pre-training model and ``MLA'' as the decoder.

\begin{table}[]
  \centering
  \caption{Accuracy of estimated semantic segmentation maps}
  \label{tab:result_cityscapes}
  \begin{tabular}{c|c|c|ccc}
  \hline
  backbone  & decoder & method       & aAcc          & mIoU           & mAcc           \\ \hline
  \multirow{9}{*}{ViT-base}
            & \multirow{3}{*}{MLA}   & baseline      & 94.74          & 71.79          & 81.50          \\ \cline{3-6}
            &                        & conventional 2 & 89.09          & 39.87          & 46.61          \\ \cline{3-6}
            &                        & proposed      & \textbf{94.54} & \textbf{70.99} & \textbf{80.85} \\ \cline{2-6}
            & \multirow{3}{*}{PUP}   & baseline      & 93.90          & 64.35          & 74.68          \\ \cline{3-6}
            &                        & conventional 2 & 89.40          & 44.65          & 54.44          \\ \cline{3-6}
            &                        & proposed      & \textbf{94.06} & \textbf{67.08} & \textbf{75.08} \\ \cline{2-6}
            & \multirow{3}{*}{Naive} & baseline      & 93.79          & 68.14          & 78.39          \\ \cline{3-6}
            &                        & conventional 2 & 88.78          & 40.59          & 49.94          \\ \cline{3-6}
            &                        & proposed      & \textbf{94.52} & \textbf{70.64} & \textbf{80.80} \\ \hline
  \multirow{9}{*}{ViT-Large}
            & \multirow{3}{*}{MLA}   & baseline      & 95.20          & 73.26          & 81.43          \\ \cline{3-6}
            &                        & conventional 2 & 92.33          & 53.55          & 62.32          \\ \cline{3-6}
            &                        & proposed      & \textbf{94.88} & \textbf{72.57} & \textbf{80.69} \\ \cline{2-6}
            & \multirow{3}{*}{PUP}   & baseline      & 94.66          & 67.14          & 76.13          \\ \cline{3-6}
            &                        & conventional 2 & 89.25          & 38.59          & 44.63          \\ \cline{3-6}
            &                        & proposed      & \textbf{94.52} & \textbf{66.20} & \textbf{74.57} \\ \cline{2-6}
            & \multirow{3}{*}{Naive} & baseline      & 94.96          & 71.38          & 81.03          \\ \cline{3-6}
            &                        & conventional 2 & 89.60          & 42.03          & 49.06          \\ \cline{3-6}
            &                        & proposed      & \textbf{94.96} & \textbf{70.51} & \textbf{79.02} \\
  \hline
  \end{tabular}
\end{table}

With the baseline, models were fine-tuned with images without encryption and query images without encryption were tested on the models.
With conventional 2, the patch embedding of the pre-train model without any encryption was fine-tuned by using encrypted images and encrypted query images were input to the models.

It was confirmed that the models trained with the proposed method performed comparable to the baseline for all decoders.
In contrast, conventional 2, which does not apply encryption to the pre-trained model (see \reffig{fig:setr_test_encryption}), showed a decrease in accuracy due to encryption.

\section{Conclusion}
We proposed a novel method for semantic segmentation that enables model training with images encrypted on cloud servers.
Using a domain-adaptation technique, which is for fine-tuning pre-train models by using encrypted images,
our method not only to protected sensitive visual information but also achieved accuracy comparable to that without any encryption when using SETR.

For future work, it is necessary to control the ability to protect visual information and to evaluate its defensive performance against various types of attacks.
We also need to do the same for ViT-based models other than SETR.

\section*{ACKNOWLEDGEMENTS}
This work was supported in part by JSPS KAKENHI (Grant Number 25K07750).


\begin{thebibliography}{99}
  \bibitem{sirichotedumrong2018grayscale}
  W.Sirichotedumrong, T.Chuman, S.Imaizumi, and H.Kiya,
  \newblock ``Grayscale-Based Block Scrambling Image Encryption for Social Networking Services,''
  \newblock in {\em 2018 IEEE International Conference on Multimedia and Expo (ICME)}, pp.1--6, 2018.

  \bibitem{kiya2022overview}
  H.Kiya, A.P.M.Maung, Y.Kinoshita, S.Imaizumi, S.Shiota, et al.,
  \newblock ``An overview of compressible and learnable image transformation with secret key and its applications,''
  \newblock {\em APSIPA Transactions on Signal and Information Processing}, vol.11, no.1, article e18, 2022.

  \bibitem{iijima2024random}
  R.Iijima, S.Shiota, and H.Kiya,
  \newblock ``A random ensemble of encrypted vision transformers for adversarially robust defense,''
  \newblock {\em IEEE Access}, 2024.

  \bibitem{aprilpyone2021block}
  M.AprilPyone and H.Kiya,
  \newblock ``Block-wise image transformation with secret key for adversarially robust defense,''
  \newblock {\em IEEE Transactions on Information Forensics and Security}, vol.16, pp.2709--2723, 2021.

  \bibitem{maungmaung2021protection}
  A.MaungMaung and H.Kiya,
  \newblock ``A protection method of trained CNN model with a secret key from unauthorized access,''
  \newblock {\em APSIPA Transactions on Signal and Information Processing}, vol.10, article e10, 2021.

  \bibitem{iida2020privacy}
  K.Iida and H.Kiya,
  \newblock ``Privacy-Preserving Content-Based Image Retrieval Using Compressible Encrypted Images,''
  \newblock {\em IEEE Access}, vol.8, pp.200038--200050, 2020.

  \bibitem{nagamori2024efficient}
  T.Nagamori, S.Shiota, and H.Kiya,
  \newblock ``Efficient fine-tuning with domain adaptation for privacy-preserving vision transformer,''
  \newblock {\em APSIPA Transactions on Signal and Information Processing}, vol.13, no.1, article e8, 2024.

  \bibitem{Zheng_2021_CVPR}
  Sixiao Zheng, Jiachen Lu, Hengshuang Zhao, Xiatian Zhu, Zekun Luo, Yabiao Wang, Yanwei Fu, Jianfeng Feng, Tao Xiang, Philip H.S. Torr, and Li Zhang.
  \newblock ``Rethinking Semantic Segmentation From a Sequence-to-Sequence Perspective With Transformers.''
  \newblock In Proceedings of the IEEE/CVF Conference on Computer Vision and Pattern Recognition (CVPR), pages 6881--6890, June 2021.

  \bibitem{dosovitskiy2021an}
  A.Dosovitskiy, L.Beyer, A.Kolesnikov, D.Weissenborn, X.Zhai, T.Unterthiner, M.Dehghani, M.Minderer, G.Heigold, S.Gelly, J.Uszkoreit, and N.Houlsby,
  \newblock ``An Image is Worth 16x16 Words: Transformers for Image Recognition at Scale,''
  \newblock in {\em Proceedings of the International Conference on Learning Representations (ICLR)}, 2021.

  \bibitem{kiya2023image}
  H.Kiya, R.Iijima, A.Maungmaung, and Y.Kinoshita.
  \newblock Image and model transformation with secret key for vision transformer.
  \newblock \textit{IEICE Transactions on Information and Systems},
  vol.106, no.1, pp.2--11, 2023.

  \bibitem{9760030}
  A.MaungMaung and H.Kiya.
  \newblock Privacy-preserving image classification using an isotropic network.
  \newblock \textit{IEEE MultiMedia}, vol.29, no.2, pp.23--33, 2022.

  \bibitem{lin2024privacy}
  H.Lin, S.Imaizumi, and H.Kiya.
  \newblock Privacy-preserving ConvMixer without any accuracy degradation using compressible encrypted images.
  \newblock \textit{Information}, vol.15, no.11, p.723, 2024.

  \bibitem{kiya2022privacy}
  H.Kiya, T.Nagamori, S.Imaizumi, and S.Shiota.
  \newblock Privacy-preserving semantic segmentation using vision transformer.
  \newblock \textit{Journal of Imaging}, vol.8, no.9, p.233, 2022.

\end{thebibliography}
\end{document}